\documentclass[letterpaper, 10 pt, conference]{ieeeconf}  

\IEEEoverridecommandlockouts                              

\usepackage{latexsym} 
\usepackage{amsmath}
\usepackage{amsthm} 
\usepackage{amsfonts,amssymb} 
\usepackage[linesnumbered, ruled, vlined]{algorithm2e}
\usepackage[percent]{overpic}
\usepackage{threeparttable} 
\usepackage{epstopdf}
\usepackage{setspace} 
\usepackage{multirow}
\usepackage{ragged2e} 
\usepackage{array} 
\usepackage{comment} 
\usepackage{cite} 
\usepackage{fancyhdr}            

\newtheorem{proposition}{Proposition}
\theoremstyle{definition}
\newtheorem{problem}{Problem}
\newtheorem{definition}{Definition}
\newtheorem{remark}{Remark}
\newtheorem*{Proof}{Proof}

\usepackage{xcolor}

\newcommand{\W}{\mathcal{W}}

\newcommand{\N}{\mathcal{N}}

\newcommand{\AP}{\mathcal{AP}}

\title{\LARGE \bf
Multi-Robot Task Planning under Individual and Collaborative Temporal Logic Specifications
}

\author{Ruofei Bai$^{1}$, Ronghao Zheng$^{1, 2, \dagger}$, Meiqin Liu$^{1, 2}$, and Senlin Zhang$^{1, 2}$
\thanks{
The authors are with the $^{1}$College of Electrical Engineering and $^{2}$State Key Laboratory of Industrial Control Technology, Zhejiang University, Hangzhou, 310027, China \{{\tt\small brf, rzheng, liumeiqin, slzhang}\}{\tt\small @zju.edu.cn}}
\thanks{$^\dagger$ Corresponding author}
}

\begin{document}

\maketitle

\thispagestyle{fancy}            
\fancyhead{}                     
\lfoot{\justifying \footnotesize \noindent \copyright 2021 IEEE. Personal use of this material is permitted. Permission from IEEE must be obtained for all other uses, in any current or future media, including reprinting/ republishing this material for advertising or promotional purposes, creating new collective works, for resale or redistribution to servers or lists, or reuse of any copyrighted component of this work in other works.}                 
\cfoot{\quad}                    

\renewcommand{\headrulewidth}{0pt}      
\renewcommand{\footrulewidth}{0pt}

\pagestyle{empty}                


\begin{abstract}
This paper investigates the task coordination of multi-robot where each robot has a private individual temporal logic task specification; and also has to jointly satisfy a globally given collaborative temporal logic task specification.
To efficiently generate feasible and optimized task execution plans for the robots, we propose a hierarchical multi-robot temporal task planning framework, in which a central server allocates the collaborative tasks to the robots, and then individual robots can independently synthesize their task execution plans in a decentralized manner.
Furthermore, we propose an execution plan adjusting mechanism that allows the robots to iteratively modify their execution plans via privacy-preserved inter-agent communication, to improve the expected actual execution performance by reducing waiting time in collaborations for the robots.
The correctness and efficiency of the proposed method are analyzed and also verified by extensive simulation experiments.

\end{abstract}

\section{INTRODUCTION}

Multi-robot task planning widely exists in many areas, such as smart logistics, autonomous inspection, intelligent manufacturing, etc. 
Planning methods based on model checking theories, such as linear temporal logic (LTL), have drawn much attention in recently years~\cite{fainekosTemporalLogicMotion2005,smithOptimalPathPlanning2011, plakuMotionPlanningTemporallogic2015,T*-A-Heuristic-ICRA-2020}, due to the user-friendly syntax of formal languages and expressive power for describing complex task requirements.
Temporal logic task planning algorithms can formally synthesize control and communication strategies for the robots by searching on a constructed automaton that captures both the task requirements and robots' capabilities.

This paper focuses on a common situation where each robot locally has its private individual LTL task specification, and all the robots also have to jointly satisfy a global collaborative LTL task specification. 
The completion of each collaborative task may require several robots of different types.
Prior studies of multi-robot task coordination under local temporal tasks usually assume that the collaboration requirements are integrated into local task specifications, i.e., assignments of collaborative tasks are fully or partially known in prior.
The execution plans are typically synthesized offline and then executed online by the robots while also reacting to environment changes.
However, little attention has been paid to reduce the waiting time for the robots in each collaboration to improve the efficiency of actual execution. 
Additionally, the concern of preserving individual robots' privacy also makes it difficult to optimize the overall performance of the execution plans.

To mitigate the above issues, we propose a hierarchical multi-robot temporal task planning framework to synthesize correct and optimized task execution plans for the robots.
The hierarchical framework involves a central server to conduct the task allocation procedure; and then the robots implement the planning and optimization procedures in a decentralized manner.
Note that we do not assume the allocation of the collaborative tasks are known in prior.
Furthermore, to reduce the waiting time in collaborations for the robots, we propose an execution plan adjusting mechanism, in which the robots can iteratively modify their execution plans to improve the expected actual performance via inter-agent communication, while also preserving the private information about robots' individual tasks.
The contributions of the paper are summarized as follows: 
\begin{enumerate}
 \item we propose a hierarchical multi-robot temporal task planning framework to efficiently synthesize execution plans for the robots satisfying both individual and global collaborative LTL task specifications, without assuming that assignments of collaborative tasks are explicitly given;
 \item Moreover, we propose an execution plan adjusting mechanism, in which the robots can iteratively modify their execution plans to improve the expected execution performance via inter-agent communication, while also preserving the private information about individual tasks;
 \item we analyze the complexity and non-optimality of the proposed method. Extensive simulation experiments verify the correctness and efficiency of the proposed method.
\end{enumerate}

\subsection{Related Works}
Existing studies in multi-robot task planning under temporal logic constraints fall into two categories: the top-down and the bottom-up pattern.
Studies in top-down pattern usually assume a globally given temporal logic specification for a team of robots~\cite{ulusoyOptimalityRobustnessMultiRobot2013}, and search for a path on a constructed product automaton that combines all robots' environment models and an automaton corresponding to the LTL formula.
Several approaches have been proposed to improve the scalability of the traditional method, such as redundant states pruning~\cite{T*-A-Heuristic-ICRA-2020, kantarosIntermittentConnectivityControl2015};
 sampling-based searching~\cite{vasileSamplingBasedTemporalLogic2013, kantarosSamplingbasedControlSynthesis2017,stylus-IJRR-2020}; and the decomposition of task specification~\cite{schillingerDecompositionFiniteLTL2018, schillingerSimultaneousTaskAllocation2018}, etc.
 
 More related to this paper, studies in bottom-up pattern typically distribute the task specification to individual robots. 
 M.~Guo~$et\ al.$~\cite{guoBottomupMotionTask2015} investigated the task coordination of loosely coupled multi-agent systems with dependent local tasks.
 The robots synthesize an off-line initial plan independently, and then execute the collaborative actions online with the assistance of other robots after the request-reply message exchanging and computation. 
 The above method is further modified in \cite{guoTaskMotionCoordination2017} to permit heterogeneous capabilities of robots and include an online task swapping mechanism. 
 J.~Tumova~$et\ al.$~\cite{tumovaDecompositionMultiagentPlanning2015} considered a slightly different setting from~\cite{guoBottomupMotionTask2015}, in which each robot have its local complex temporal logic specifications,  including an independent motion specification; and a dependent collaborative task specification.
 An two-phase automata-based method was proposed, where each robot synthesizes its motion planning on a local automaton; then a centralized planning procedure is conducted on a product automaton of the pruned local automaton where only states related to the collaborative tasks are left.
 Despite the size of the product automaton have been greatly reduced by the pruning techniques, the method still has exponential complexity.
 
Y. Kantaros~$et\ al.$~\cite{kantarosTemporalLogicTask2019} analyzed a similar situation to this paper, that robots have their independent individual task specifications, while they have to jointly satisfy the global intermittent communication tasks, that require several robots to intermittently gather in some locations to exchange information.
The paper proposes an algorithm in which the robots iteratively optimize their gathering locations to reduce the total traveling distance via on-line communication.

Inspired by~\cite{kantarosTemporalLogicTask2019}, we extend the global collaborative task specification into finite LTL; and aim to minimize the total time spent to finish all tasks considering reducing the waiting time for the robots in each collaboration, which has not been considered in~\cite{kantarosTemporalLogicTask2019}.
Furthermore, the planning and optimization procedures are all performed in decentralized manner, so that robots' private information about their individual tasks can be preserved.

\section{Preliminaries and Problem Formulation}
\subsection{Linear Temporal Logic}
An LTL formula $\varphi$ over a set $\mathcal{AP}$ of atom propositions are defined according to the following recursive grammar~\cite{baierPrinciplesModelChecking2008}:
\[\varphi::={\rm true}\ |\ \pi\ |\ \varphi_{1}\wedge \varphi_{2}\ |\ \neg \varphi\ |\ \bigcirc \varphi\ |\ \varphi_{1} \rm U \varphi_{2},\] 
where $\mathrm{true}$ is a predicate true and $\pi\in \mathcal{AP}$ is an atom proposition. 
The other two frequently used temporal operators $\Diamond$~(eventually) and $\Box$~(always) can be derived from operator ${\rm U}$~(until). 
In this paper, we only consider a kind of finite LTL~\cite{giacomoReasoningLTLFinite} called ${\rm LTL}_{f}$, which is a variant of original LTL that can be interpreted over finite sequences, and uses the same syntax as original LTL.
Moreover, we exclude the $\bigcirc$ (next) operator from the syntax, since it is not meaningful in practical applications~\cite{a-fully-automated-TAC-2008}.
We refer the reader to \cite{baierPrinciplesModelChecking2008} for the details of LTL semantics.

Given an ${\rm LTL}_{f}$ formula $\varphi$, a nondeterministic finite automaton (NFA) can be constructed which accepts exactly the sequences that make $\varphi$ true.

\begin{definition}[Nondeterministic Finite Automaton]
	A nondeterministic finite automaton (NFA) $\mathcal{F}$ is a tuple $\mathcal{F} := \langle \mathcal{Q}_{F},\mathcal{Q}_{F}^{0}, \alpha, \delta, \mathcal{Q}_{F}^{F}\rangle$, where 
	$\mathcal{Q}_{F}$ is a finite set of states;
	$\mathcal{Q}_{F}^{0}\subseteq \mathcal{Q}_{F}$ is a set of initial states;
	$\alpha$ is a set of Boolean formulas over $\pi \in \AP$;
	$\delta:\mathcal{Q}_{F}\times \mathcal{Q}_{F}\rightarrow \alpha$ is the transition condition of states in $\mathcal{Q}_{F}$; 
	$\mathcal{Q}_{F}^{F}$ is a set of accepting states.
\end{definition}

A finite sequence of states $q_{F}\in \mathcal{Q}_{F}$ is called a run $\rho_{F}$. A run $\rho_F$ is \emph{accepting} if it starts from one initial state $q_{F}^{0}\in \mathcal{Q}_{F}^{0}$ and ends at an accepting state $q_{F}^{F}\in \mathcal{Q}_{F}^{F}$. 
Given a finite sequence $\sigma = \sigma(1)\sigma(2)\dots\sigma(L-1)\sigma(L)$, we say that $\sigma$ describes a run $\rho_F$ if $\sigma(i)\vDash \delta\left(\rho_{F}(i), \rho_{F}(i+1)\right)$ for all $i\in \left[1:L-1\right]$, i.e., $\sigma(i)$ is a set of atom propositions or negative propositions, which makes the Boolean formula $\delta\left(\rho_{F}(i), \rho_{F}(i+1)\right)$ become ${\rm true}$.
Here $\left[1:L-1\right]$ denotes a set of indexes increasing from $1$ to $L-1$ by step $1$.
A finite sequence $\sigma$ fulfills an ${\rm LTL}_{f}$ formula if at least one of its runs is accepting.
The minimum requirements for a sequence to describe a run can be represented as an essential sequence, as in the following definition.
\begin{definition}[Essential Sequence]
\label{essential-sequence}
Considering a sequence $\sigma = \sigma(1)\dots\sigma(L)$, it is called \emph{essential} if and only if it describes a run $\rho_F$ in NFA $\mathcal{F}$ and $\sigma(i)\backslash \{\pi\} 
\nvDash \delta(\rho_{F}(i), \rho_{F}(i+1))$ for all $\pi\in \sigma(i)$ and $i\in \left[1:L-1\right]$.
\end{definition}

\begin{figure}[t]
	\flushleft
	\begin{overpic}
		[scale=0.125]{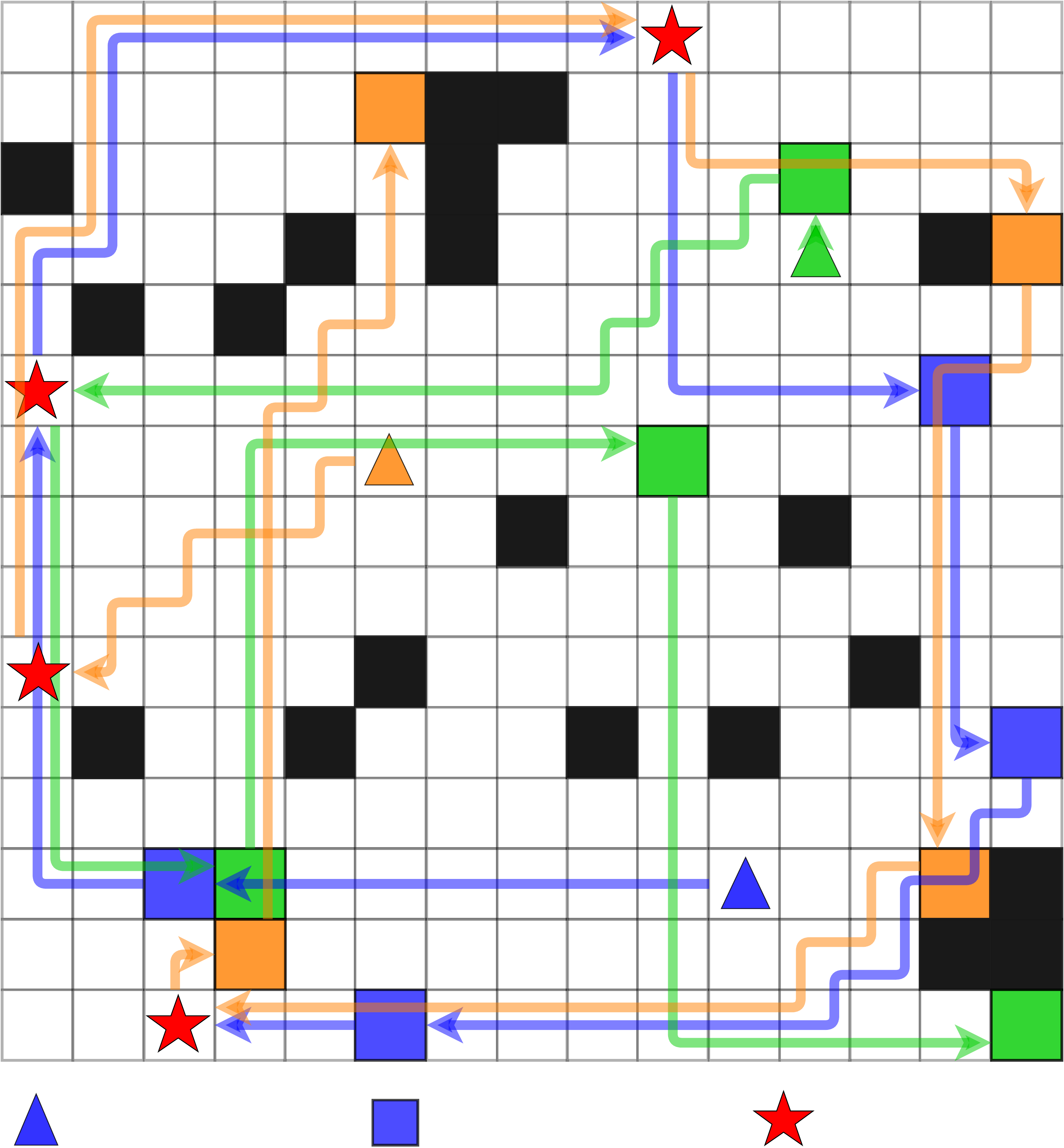}
        \put(100, 85){\scriptsize $\varphi_{1} = (\Diamond \pi^{ts_{1}}) \wedge (\Diamond \pi^{ts_{2}}) \wedge (\Diamond \pi^{ts_{3}}) \wedge$}
        \put(117, 75){\scriptsize $ (\Diamond \pi^{ts_{4}}) \wedge (\neg \pi^{ts_{1}} ~{\rm U}~ \pi^{ts_{4}})$}
        \put(100, 65){\scriptsize $\varphi_{2} = \cdots$}
        \put(100, 55){\scriptsize $\dots$}
        \put(100, 25){\scriptsize $\phi = (\Diamond \pi^{ct_{1}}) \wedge (\Diamond \pi^{ct_{2}}) \wedge (\Diamond \pi^{ct_{4}}) \wedge $}
        \put(113,15){\scriptsize $(\neg \pi^{ct_{3}}~{\rm U}~\pi^{ct_{2}})  \wedge$}
        \put(113,5){\scriptsize $(\Box(\pi^{ct_{4}} \rightarrow (\Diamond \pi^{ct_{3}})))$}
        \put(8,0){\scriptsize robot $i$}
        \put(40,0){\scriptsize $ts\in \overline{\mathbb{T}}_i$}
        \put(74,0){\scriptsize $ct\in \widetilde{\mathbb{T}}$}
	\end{overpic}
	\caption{The simulation results of an experiment with $3$ robot.}
	\label{environment}
\end{figure}
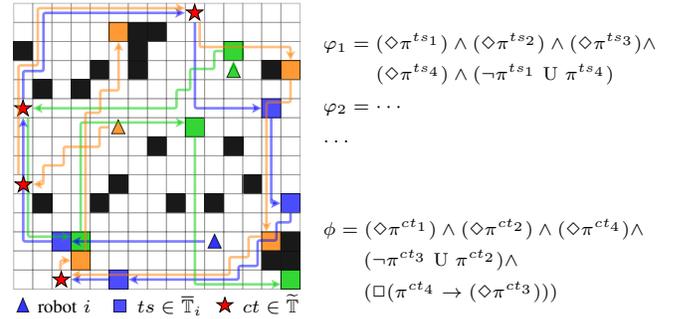

\subsection{System Model}
\label{ts_ltl}

Consider there is a set of robots $\mathcal{N}:=\{1, ..., N\}$ operating in an environment, e.g., as in Fig.~\ref{environment}, which can be represented as a graph $\mathcal{W} = \langle\mathcal{Q}, \mathcal{E}\rangle$.
Here $\mathcal{Q}$ is the set of regions of interest and $\mathcal{E}$ contains the connectivity relations of regions in $\mathcal{Q}$. 
Each robot $i\in \mathcal{N}$ has its specific capability $c_j\in Cap$ and can provide service $c_j$ in the environment $\mathcal{W}$. 
The set $Cap:=\{c_j\}_{j\in \{1, ..., |Cap|\}}$ contains all capability types.
We define $\N_j:=\left\{i~|\text{robot $i$ has the capability $c_j$}\right\}$.

There are several tasks distributed in $\W$ to be completed by the robots, as defined in Definition~\ref{def-1}.

\begin{definition}[Task Requirement]
\label{def-1}
A task in $\mathcal{W}$ is a tuple $ts := \langle \pi^{ts}, \{(c_j, m_j^{ts})\}_{j\in I_{ts}}, q^{ts}\rangle$, where $\pi^{ts}$ is the unique atom proposition of $ts$; $q^{ts}\in \mathcal{Q}$ corresponds to the region associated with $ts$; and $ts$ can be completed, i.e, $\pi^{ts}$ becomes $\mathrm{true}$, if at least $m_{j}^{ts}$ robots with capability $c_{j}$ are deployed simultaneously for all $j\in I_{ts}$ and perform actions according to their capabilities, where
$I_{ts}\subseteq \{1, ..., |Cap|\}$ contains indexes of capabilities required by $ts$.
\end{definition}

In this paper, we consider the situation that: (1) each robot $i\in \mathcal{N}$ has a pre-assigned private individual task specifications $\varphi_i$ that can be satisfied by itself. 
Here $\varphi_i$ is an ${\rm LTL}_{f}$ formula defined over
$\overline{\mathbb{T}}_i$, which is robot $i$'s individual task set.
For each $ts\in \overline{\mathbb{T}}_i$, it holds that $|I_{ts}|=1$, $m^{ts}_j=1$, and $i\in \mathcal{N}_j$;
(2) the robots have to jointly satisfy a global ${\rm LTL}_{f}$ task specification $\phi$ defined over a set $\widetilde{\mathbb{T}}$ of tasks, in which each task $ct\in \widetilde{\mathbb{T}}$ may has much heavier workloads and requires no less than one robot with several different capabilities.
Note that we particularly use $ct$ to represent a task in $\widetilde{\mathbb{T}}$, called \emph{collaborative task}, to distinguish it from tasks in the set $\bigcup_{i\in \mathcal{N}}\overline{\mathbb{T}}_i$ of all \emph{individual tasks} in the following context.


We assume that the execution of each individual task specification $\varphi_{i}$ is independent of each other, i.e., does not influence the states of other robots. 
And they are also independent of the execution of collaborative tasks.
More formally, we assume that: $\forall~i, j\in \mathcal{N}$ and $i\ne j$, $\overline{\mathbb{T}}_i\cap \overline{\mathbb{T}}_j=\emptyset$; and $\forall~i\in\mathcal{N}$, $\widetilde{\mathbb{T}}\cap \overline{\mathbb{T}}_i=\emptyset$.
Otherwise, the non-independent task requirements can be formulated into the global collaborative task specification $\phi$. 

The mobility and capability of each robot $i$ can be formalized as a weighted transition system.

\begin{definition}[Weighted Transition System]
A weighted transition system of robot $i$ is a tuple $wTS_{i} = \langle \mathcal{Q}_i, q_{i}^{0}, \\ \rightarrow_{i}, \omega_{i}, \mathcal{AP}_{i}, L_{i} \rangle$, where $\mathcal{Q}_{i}\subseteq \mathcal{Q}$ is a finite set of states corresponding to regions in $\mathcal{W}$; $q_i^0$ is the initial state; $\rightarrow_{i}\subseteq \mathcal{E}$ contains all pairs of adjacent states; $\omega_{i}: \mathcal{Q}_{i}\times \mathcal{Q}_{i} \rightarrow \mathbb{R}^{+}$ is a weight function measuring the cost for each transition; $\mathcal{AP}_{i}$ is the set of atom propositions related to the tasks in $\mathcal{W}$; $L_{i}: \mathcal{Q}_{i} \rightarrow 2^{\mathcal{AP}_{i}}$ is a labeling function, 
and satisfies that $(1)$ $\forall~ts\in \overline{\mathbb{T}}_i$, $\pi^{ts}\in L_{i}(q^{ts})$; $(2)$ $\forall~ct\in \widetilde{\mathbb{T}}$, $\pi^{ct}\in L_{i}(q^{ct})$ iff $\exists j\in I_{ct}$, $i\in \mathcal{N}_j$. Here $q^{ts}, q^{ct}\in \mathcal{Q}_{i}$ are the states related to the individual task $ts$ and collaborative task $ct$ respectively.
\end{definition}


\subsection{Problem Formulation}
Let $T^{\rm colla}$ denotes the total time cost for all robots to satisfy all task requirements. 
$T^{\rm colla} = \sum_{i\in \mathcal{N}} T^{\rm colla}_{i}$, where $T^{\rm colla}_{i}$ denotes the amount of time robot $i$ spends to execute $\tau_i$, considering the wait time in each collaboration (due to the different arrival times of the robots in the collaboration). 

\begin{problem}
\label{problem1}
Consider a set $\mathcal{N}$ of robots operate in an environment  $\mathcal{W}$, and suppose that the mobility and capability of each robot $i$ is modeled as a weighted transition system $wTS_{i}$. 
Given individual ${\rm LTL}_{f}$ task specification $\varphi_{i}$ for each robot $i$ and a global collaborative task specification $\phi$, find task execution plan $\tau_{i}$ in $wTS_{i}$ for each robot $i$ which satisfies:
\begin{enumerate}
    \item  the execution of $\tau_{i}$ satisfies individual task specification $\varphi_{i}$;
    \item the joint behaviors of all robots satisfies collaborative task specification $\phi$; 
    \item the total time cost $T^{\rm colla}$ is minimized.
\end{enumerate}
\end{problem}

\section{Multi-Robot Temporal Task planning}

\begin{figure}[!t]
	\centering
	\begin{overpic}
		[scale=0.16]{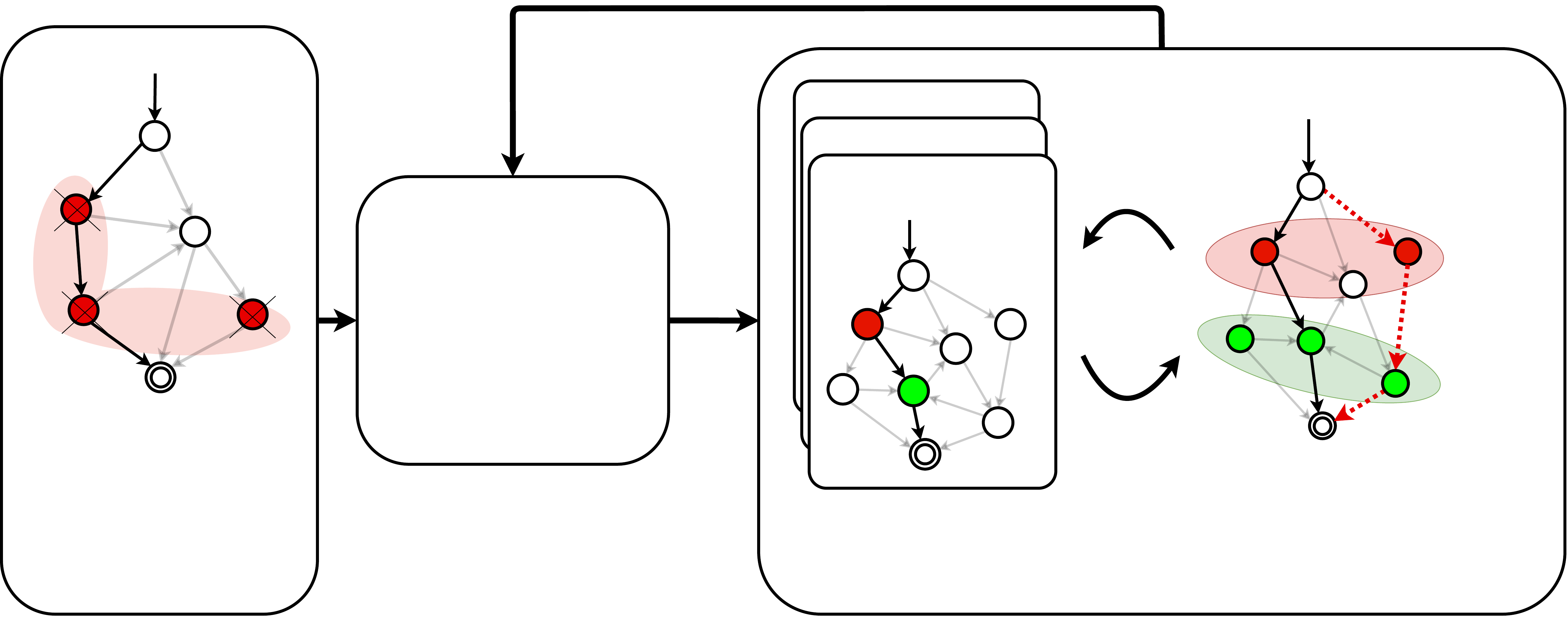}
		\put(5.5,5){\scriptsize Sequence}
		\put(5.5,2){\scriptsize Selection}
		\put(23.5,23){\scriptsize Z3 SMT solver}
		\put(23.5,18){\scriptsize $f \leftarrow f \wedge \neg X$}
		\put(30.5,5){\scriptsize Task}
		\put(28,2){\scriptsize Allocation}
		\put(57,5){\scriptsize Local}
		\put(55,2){\scriptsize Planning}
		\put(77,5){\scriptsize Execution Plan}
		\put(80,2){\scriptsize Adjusting}
		\put(54,27.5){\tiny Robot $1$: $\mathcal{P}_{1}$}
		\put(53,30.35){\tiny Robot $2$: $\mathcal{P}_{2}$}
		\put(52,32.5){\tiny Robot ...}
		\put(44,21){\scriptsize $\mathcal{X}$}
		\put(4,11){\tiny NFA $\mathcal{F}$ of $\phi$}
		\put(90,26){\tiny $\mathcal{C}(ct^{k}_{l})$}
		\put(90,12){\tiny $\mathcal{C}(ct^{k'}_{l'})$}
		
	\end{overpic}
	\caption{Framework of multi-robot temporal task planning.}
	\label{framework}
\end{figure}
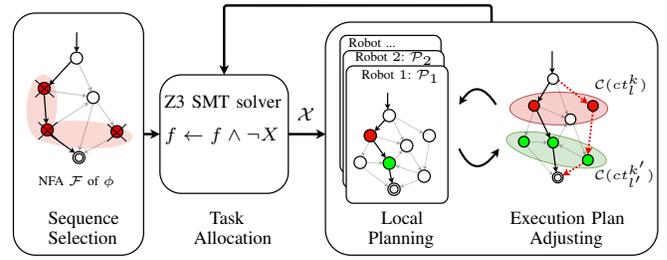

 In this section, we propose the multi-robot temporal task planning framework to solve Problem~\ref{problem1}, as shown in Fig.~\ref{framework}. 
 The proposed framework firstly finds a sequence of collaborative tasks which satisfies the collaborative task specification and allocates the tasks in the sequence to robots. 
 Then each robot locally synthesizes its initial task execution plan which satisfies the individual task specification and the assigned collaborative tasks requirements. 
The initially obtained execution plans will be further improved by a proposed execution plan adjusting mechanism, which will be explained in Sect.~\ref{section-greedy}. 
 The detailed steps of the framework are described in Alg.~\ref{alg-framework}.
 
 \subsection{Selection and Decomposition of Execution Sequence}
 \label{3A}
 To find a collaborative task sequence satisfying the collaborative task specification $\phi$, a three-step
 scheme is developed:

  \begin{enumerate}
      \item prune the NFA $\mathcal{F}$ of $\phi$ by removing all impossible transitions that require assistance beyond what the robot set $\mathcal{N}$ can provide.
      \item identify the decomposition states in the pruned $\mathcal{F}$ by utilizing the decomposition algorithm proposed in~\cite{schillingerDecompositionFiniteLTL2018}.
      \item select an essential sequence which describes an accepting run in $\mathcal{F}$, and divide it into independent subsequences by the decomposition states along the run. 
  \end{enumerate}

The way to prune $\mathcal{F}$ is straightforward and we omit the details here. 
The essential sequence $\sigma$ that describes an accepting run $\rho_{F}$ in the pruned NFA $\mathcal{F}$ is actually a collaborative task sequence that satisfies $\phi$.
Given an accepting run $\rho_{F}$ of $\mathcal{F}$ and its corresponding essential sequence $\sigma$, the \emph{decomposition states} along the run $\rho_{F}$ can decompose $\sigma$ into $S$ $(\ge 1)$ subsequences, i.e., $\sigma = \sigma^{1} ;\sigma^{2};\dots;\sigma^{S}$.
Let $\mathcal{S}:=\left[1:S\right]$ denote the set of indexes.
Such decomposition is proved to have two properties in \cite{schillingerSimultaneousTaskAllocation2018}: (1) \textbf{Independence}. Execution of each subsequence $\sigma^{k}$ will not violate another $\sigma^{j}$, $\forall~k, j\in \mathcal{S}$ and $k\ne j$; (2) \textbf{Completeness}. The completion of all subsequences $\sigma^{k}$ implies the completion of $\sigma$.
The detailed definition of decomposition states can be found in Definition~$9$ and Theorem $2$ of \cite{schillingerDecompositionFiniteLTL2018}.

We define $\sigma^{k}(m) := \{ct^{k}_{(m, j)}\}_{j\in \left[1:|\sigma^{k}(m)|\right]}$ to denote the $m$-th element in $\sigma^{k}$, in which $ct^{k}_{(m,j)}\in \widetilde{\mathbb{T}}$ denotes the $j$-th task in the $m$-th element of $\sigma^{k}$.
For notation simplicity, we use $ct$ and $\pi^{ct}$ interchangeably afterward.
Note that we only consider the positive atom propositions in the essential sequence $\sigma$, and assume that the negative propositions are guaranteed in actual execution as safety guards, which may impose implicit communications in actual deployment.

The execution of collaborative tasks within each $\sigma^{k}$ must satisfy the following temporal constraints according to the accepting condition of $\mathcal{F}$.
\begin{itemize}
    \item  \textbf{Synchronization Constraints}. ${\forall}\ ct^{k}_{(m, j)}$, 
    $ct^{k}_{(m, j')}\in \sigma^{k}(m)$, $j\ne j'$, 
    $ct^{k}_{(m, j)}$ and $ct^{k}_{(m, j')}$ need to be executed synchronously;
    \item   \textbf{Ordering Constraints}. $\forall~ct^{k}_{(m, j)}$, $ct^{k}_{(m', j')}\in \sigma^{k}, m<m'$, 
    $ct^{k}_{(m', j')}$ cannot be executed before $ct^{k}_{(m, j)}$.
\end{itemize}

The above decomposition of essential sequence relaxes the temporal constraints between the tasks in the essential sequence, which may help to reduce the unnecessary communication as well as the total time cost in actual deployment.


The selection of the collaborative task sequence $\sigma$ has implicit influences on the performance of the final results.
To reduce the complexity, in the sequel we just select the essential sequence $\sigma$ that describes one of the shortest accepting run $\rho_{F}$ in the pruned $\mathcal{F}$ to be the collaborative task sequence. 
Simulation results illustrate that the proposed method can find solutions of high quality in practice under the selected~$\sigma$.

\subsection{SMT-Based Collaborative Task Allocation}
\label{3B}

After obtaining the decomposed collaborative task sequence $\sigma = \sigma^{1} ;\sigma^{2};\dots;\sigma^{S}$ and the related accepting run $\rho_{F}$ of $\mathcal{F}$, we now consider the allocation of collaborative tasks in the sequence $\sigma$. 

For notation simplicity, we give all $ct^{k}_{(m,j)}\in \sigma^{k}$ an index $l $, defined as $ct^{k}_{l}:=ct^{k}_{(m, j)}$, where $l = \sum_{i = 1}^{m-1} |\sigma^{k}(i)| + j$, $\forall~m\in \left[1:|\sigma^{k}|\right]$ and $k\in \mathcal{S}$.
Then the $l$-th task in $\sigma^{k}$ can be referred as $ct^{k}_{l}$.
We define a set of Boolean variables $\mathcal{X}_{i} := \{
x_{i}^{(k, l)}|ct^{k}_{l}\in \sigma^{k}, k\in \mathcal{S}
\}$ for each robot $i$ to indicate the task assignment results. 
A true $x_{i}^{(k, l)}$ implies that robot $i$ is assigned to complete the task $ct^{k}_{l}$.
Now we construct the SMT (Satisfiability Modulo Theories) model of the task allocation problem. 

(1) $\textbf{Collaboration Requirements.}$ A feasible task assignment must satisfy the amount and types of assistance needed to complete each collaborative task, as the following constraints:
for each $\sigma^k$,
${\forall}\ ct^{k}_{l}\in \sigma^{k}$,
\[
\sum_{i\in \mathcal{N}} \mathbf{1}\left(x_{i}^{(k, l)} \bigwedge i\in \mathcal{N}_{j}\right ) \ge m_{j}^{ct}, {\forall} j \in I_{ct} 
\]
is satisfiable, where $\mathbf{1}(\cdot)$ is an indicator function defined as $\mathbf{1}({\rm true}) = 1$ and $\mathbf{1}({\rm false}) = 0$.

(2) $\textbf{At Most One at a Time.}$ Each robot cannot participate in two distinct collaborative tasks which are executed synchronously. The constraints can be captured by: for each $\sigma^k$,
${\forall}\ ct^{k}_{l_{1}}, ct^{k}_{l_{2}}\in \sigma^{k}(m), l_{1}\ne l_{2}$,
\[
\neg x_{i}^{(k,l_{1})} \bigvee \neg x_{i}^{(k,l_{2})} , {\forall}\ i\in \mathcal{N} 
\]
is satisfiable.

(3) $\textbf{Communication Reduction.}$ The intersection of two sets of robots that complete two consecutive collaborative tasks is not empty: for each $\sigma^k$,
\[\bigvee_{i\in \mathcal{N}} \left(\vee_{l \in  \{l^{k}_{m-1}+1,..., l^{k}_{m}\}} x_{i}^{(k,l)}  \bigwedge \vee_{l \in  \{l^{k}_{m}+1,..., l^{k}_{m+1}\}} x_{i}^{(k,l)} \right)\]
is satisfiable, where $l^{k}_{m} = \sum_{j = 1}^{m}|\sigma^{k}(j)|$, $\forall~m\in [1:|\sigma^{k}|-1]$.

When the communication is limited in the environment, the above constraints $(3)$ can be applied to guarantee that each robot $i$ only needs to communicate with its assistant robots $j\in \mathcal{R}(ct)\backslash \{i\}$ when executing the assigned collaborative task $ct$.
The function $\mathcal{R}:ct^{k}_{l}\rightarrow 2^{\mathcal{N}}$ maps $ct^{k}_{l}$ to a set of robots that perform the task.
The constraint $(3)$ requires that $\mathcal{R}(\sigma^{k}(m))\bigcap \mathcal{R}(\sigma^{k}(m+1))\ne\emptyset$, $\forall~m\in \left[1:|\sigma^{k}|-1\right]$. 
The robot $i\in \mathcal{R}(\sigma^{k}(m))\bigcap \mathcal{R}(\sigma^{k}(m+1))$ behaves like a coordinator to guarantee the execution order of tasks in  $\sigma^{k}(m)$ and $\sigma^{k}(m+1)$.
Here the function $\mathcal{R}$ is reloaded as $\mathcal{R}(\sigma^{k}(m)) = \bigcup_{ct\in \sigma^{k}(m)} \mathcal{R}(ct)$ according to SMT constraint (2) and the synchronization constraints in Sect.~\ref{3A}.

\begin{remark}
    The SMT constraints $(3)$ can be selectively applied to some pairs of consecutive collaborative tasks, whose locations may be far apart so that it is difficult to ensure the connectivity of the communication links between these regions. 
\end{remark}

The overall SMT formula $f$ is a conjunction of all Boolean expressions of the above constraints $(1)\,(2)\,(3)$.
We define $X = \bigwedge_{i\in \mathcal{N}}\left(\wedge_{x_{i}^{(k,l)}\in \mathcal{X}_{i}}x_{i}^{(k,l)}\right)$ as the Boolean formula corresponds to a feasible task assignment $\{\mathcal{X}_i\}_{i\in \mathcal{N}}$ which satisfies~$f$. 
By utilizing an SMT solver, we can iterate all valid assignments by adding the negation of current $X$ into $f$, i.e., $f = f\wedge\neg X$ (Line 13 of  Alg.~\ref{alg-framework}). 
Each feasible task assignment $\{\mathcal{X}_i\}_{i\in\mathcal{N}}$ will be passed into the subsequent local planning procedure to further investigate its performance.

Here we come up with a filtering strategy to fast filter the non-optimal task assignments. The efficiency of the filtering strategy will be evaluated in Sect.~\ref{simulation}.

$\textbf{Filtering Strategy:}$ 
If ${\exists}\ m<n$ and $m, n\in \mathbb{N}^{+}$, such that ${\forall} i\in \mathcal{N}$, $\mathcal{X}^{n}_{i}\ge \mathcal{X}^{m}_{i}$, then $f$ can be directly modified as $f\wedge \neg X^{n}$ without proceeding the subsequent procedures, where $\mathcal{X}^{m}_{i}$, $\mathcal{X}^{n}_{i}$ are the feasible task assignments
for robot $i$ in $m$-th and $n$-th iterations, respectively. 
Here $\mathcal{X}^{n}_{i}\ge \mathcal{X}^{m}_{i}$ means that $\forall~^n{x}_{i}^{(k,l)}\in \mathcal{X}^{n}_{i}$ and $^{m}{x}_{i}^{(k,l)}\in \mathcal{X}^{m}_{i}$,
$\mathbf{1}(^n{x}_{i}^{(k,l)})\ge \mathbf{1}(^{m}{x}_{i}^{(k,l)})$.

Given the obtained task assignment results, each robot will combine the assigned collaborative tasks with its local task specification to synthesize its initial execution plan. 
The details will be discussed in the next subsection.

\subsection{Execution Plan Synthesis With Collaborative Tasks}
\label{3c}
To synthesize the individual task execution plan for each robot, the key problem is how to construct a local ${\rm LTL}_f$ formula that captures the individual task specification $\varphi_i$, as well as the temporal constraints of the assigned collaboration tasks.

Given one feasible task assignments $\mathcal{X}$, we construct set $\widetilde{\mathbb{T}}_{i}^{k} := \left\{ct^{k}_{l}\ \big\vert\ x_{i}^{(k, l)}~{\rm is~true}\right\}$ for all $k$ and $i$.
Let $\widetilde{\mathbb{T}}_{i} :=\bigcup_{k\in \mathcal{S}} \widetilde{\mathbb{T}}_{i}^{k}$. We sort all tasks in $\widetilde{\mathbb{T}}_{i}$ 
in the increasing order of indexes $k$ and $l$.

The modified local ${\rm LTL}_f$ formula $\widetilde{\varphi}_i$ for each robot $i$ is the conjunction of $\varphi_i$ and $\phi_i$, i.e., $\widetilde{\varphi}_i = \varphi_i \bigwedge \phi_i$. 
Here $\phi_i$ is the ${\rm LTL}_f$ formula that captures the requirements of tasks $ct\in \widetilde{\mathbb{T}}_{i}$.
There are two steps to construct $\phi_i$ and $\widetilde{\varphi}_i$:

\textbf{Step 1:} Formalize ordering constraints within each independent subsequence.  
We initialize the collaborative task specification $\phi_{i}^{k}$ corresponds to $\widetilde{\mathbb{T}}_{i}^{k}$ as $\phi_{i}^{k} = \Diamond(ct^{k}_{l_{1}})$, where $ct^{k}_{l_{1}}$ is the first task in the sorted $\widetilde{\mathbb{T}}_{i}^{k}$.
Then we use $ct^{k}_{l_{m}} \bigwedge \Diamond (ct^{k}_{l_{m+1}})$ to iteratively substitute $ct^{k}_{l_{m}}$ in $\phi_{i}^{k}$, where $ct^{k}_{l_{m}}$ and $ct^{k}_{l_{m+1}}$ are two consecutive tasks in sorted $\widetilde{\mathbb{T}}_{i}^{k}$, and $l_{m} < l_{m+1}$. 

The asynchronous execution of several independent subsequences may stick into deadlock in actual deployment, due to the different execution order in each robot's local execution plan. We use Step~$2$ to prevent the deadlock.

\textbf{Step 2:} Determine the execution order of each independent subsequence. For each robot $i$, we sort $\phi_{i}^{k}$ in Step~1 in increasing order of index $k$, $\forall~k\in \mathcal{S}$. 
Starting from $k=1$ to $S-1$, we iteratively replace $ct^{k}_{l_{max}}$ with $ct^{k}_{l_{max}} \bigwedge \Diamond(\phi^{k+1}_{i})$, where $ct^{k}_{l_{max}}$ is the last collaborative task in sorted $\widetilde{\mathbb{T}}^{k}_{i}$. Finally, $\phi_{i} = \phi_{i}^{1}$, and $\widetilde{\varphi}_{i} = \varphi_{i}\bigwedge \phi_{i}$.

Given the $wTS_i$ and $\mathcal{F}_i$ of $\widetilde{\varphi}_{i}$, each robot $i$ can find its initial individual optimal run $\rho_{i}$ by searching for the shortest accepting run on a product automaton $\mathcal{P}_{i}$, as defined in Definition~\ref{pa}. 
The initial task execution plan $\tau_i$ of robot $i$ can be obtained by projecting the run $\rho_i$ into the state space of $wTS_{i}$, i.e., $\tau_i = \Pi_{i} \rho_i$.

\begin{definition}[Product Automaton]
\label{pa}
	The product automaton of $wTS_{i}$ and $\mathcal{F}_i$ is a tuple $\mathcal{P}_i=wTS_{i}\otimes \mathcal{F}_i = \langle \mathcal{Q}_{P},\mathcal{Q}_{P}^{0},\rightarrow_{P}, \omega_{P},\mathcal{Q}_{P}^{F} \rangle$, 
	where 
	$\mathcal{Q}_{P} = \mathcal{Q}_{i}\times \mathcal{Q}_{F}$;
	$\mathcal{Q}_{P}^{0} = q^{0}_{i}\times \mathcal{Q}_{F}^{0}$;  
	$\rightarrow_{P} \subseteq \mathcal{Q}_{P} \times \mathcal{Q}_{P}$, which satisfies that $\forall~(q_{P},q_{P}')\in \rightarrow_{P}$, it holds that
	$(\Pi_{i}q_{P}, \Pi_{i}q_{P}')\in \rightarrow_{i}$ and $L_{i}(\Pi_{i} q_{P}) \vDash \delta(\Pi_{F}q_{P}, \Pi_{F}q_{P}')$.
	Here $\Pi_{i}$ and $\Pi_{F}$ represent the projections into the state spaces of $wTS_{i}$ and $\mathcal{F}_i$, respectively;
	$\omega_{P}(q_{P},q_{P}')=\omega_{i}(\Pi_{i}q_{P},\Pi_{i}q_{P}')$;
	$\mathcal{Q}_{P}^{F}\subseteq \mathcal{Q}_{i}\times \mathcal{Q}_{F}^{F}$. 
\end{definition}

In actual execution, the time instances when robots take part in each collaboration may be different, since robots synthesize their task execution plans independently without considering others.
In the next section, the initial task execution plans will be optimized.

\section{Greedy-based Execution Plan Adjustment}
\label{section-greedy}

In this section, we propose a greedy-based execution plan adjusting mechanism that can reduce the waiting time in collaborations (due to the different arrival times of robots in each collaboration) to optimize the total time cost.
We define the collaborative state to identify the states in $\mathcal{P}_i$ related to the execution of collaborative tasks $ct^{k}_{l}\in \widetilde{\mathbb{T}}_{i}$:
\begin{definition}[Collaborative State]
\label{collaborativestate}
For ${\forall}\ q_{P}\in \mathcal{Q}_{P}$, $q_{P}\in \mathcal{C}(ct^{k}_{l})$ iff the following two conditions hold. Here $\mathcal{C}(ct^{k}_{l})$ is a set of all collaborative states of collaborative task $ct^{k}_{l}$. 
\begin{enumerate}
    \item $ct^{k}_{l}\in L_{i}(\Pi_{i}q_{P})$;
    \item ${\exists} q_{P}'\in \mathcal{Q}_{P}, q_{P} \in \delta(q_{P}', ct^{k}_{l})$, and $\Pi_{\mathcal{F}}q_{P}' \ne \Pi_{\mathcal{F}}q_{P}$.
\end{enumerate}
\end{definition}

The condition $2)$ ensures that the robot plans to perform task $ct^{k}_{l}$ at state $q_{P}$ rather than just going through it.

\begin{algorithm}[!t]
\setstretch{1}
\SetKwInOut{Input}{Input}\SetKwInOut{Output}{Output}
\SetKwInOut{Return}{Return}
\label{alg-framework}
\caption{Framework}
\small{
\Input{$\{\varphi_{i}\}_{i\in \mathcal{N}}$, $\phi$,  $\mathcal{W}$}
\Output{$\{\tau_{i}\}_{i\in \mathcal{N}}$}
Construct $wTS_{i}$ for $i\in \mathcal{N}$.\\
Convert $\phi$ to its corresponding NFA $\mathcal{F}$.\\
Prune $\mathcal{F}$ and identify the decomposition states in $\mathcal{F}$. Select and decompose the collaborative task execution sequence $\sigma = \sigma^{1} ;\sigma^{2};\dots;\sigma^{S}$ of an accepting run $\rho_{F}$ in $\mathcal{F}$.\\
Construct the SMT-based task allocation model $f$.\\
\While{$\textbf{{\rm SAT}}(f)$}{
    $\mathcal{X}\leftarrow \textbf{\rm SMT-solver}(f)$\\
    \For{$i\in \mathcal{N}$}{
        $\phi_i \leftarrow$ the ${\rm LTL}_f$ formula captures the assigned $ct\in \widetilde{\mathbb{T}}_i$ and the temporal constraints\\
        $\widetilde{\varphi}_{i} \leftarrow \varphi_{i}\bigwedge \phi_{i}$         ~~~~~~~$//$ according to $\mathcal{X}$\\  
        $\mathcal{P}_{i}\leftarrow wTS_{i}\bigotimes \mathcal{F}_i$~~~~~~~$//$  $\mathcal{F}_i$ is the NFA of $\widetilde{\mathcal{\varphi}}_i$\\
        $\rho_{i}\leftarrow$ local optimal path of $\mathcal{P}_{i}$\\
    }
    $\textbf{{\rm ExecutionPlanAdjusting}}\left(\{\mathcal{P}_{i}\}_{i\in \mathcal{N}}, \{\rho_{i}\}_{i\in \mathcal{N}}\right )$\\
    $f\leftarrow f \bigwedge\neg X$\\
}
\KwRet{$\tau_{i}\leftarrow \Pi_{i}\rho_{i}$ for each robot $i\in\mathcal{N}$.}
}
\end{algorithm}

\begin{algorithm}[!t]
\setstretch{1}
\SetKwInOut{Input}{Input}\SetKwInOut{Output}{Output}
\label{timecost}
\caption{TimeCost}
\small{
 ${\rm delay}_{i} = 0$, for $i \in \mathcal{N}$.\\
 \For{$ct\in \widetilde{\mathbb{T}}^{sort}$}{
    $t(ct) = \max_{i\in \mathcal{R}(ct)}\{t_{i}(ct) + {\rm delay}_{i}\}$\\
    \For{$i\in \mathcal{R}(ct)$}{
        ${\rm delay}_{i} = t(ct) - t_{i}(ct)$\\
    }
 }
 $T_{i}^{\rm colla}= T_{i}^{\rm indiv} + {\rm delay}_{i}$, for $i\in \mathcal{N}$.\\
\KwRet{$T^{\rm colla} = \sum_{i\in \mathcal{N}}T_{i}^{\rm colla}$}
}
\end{algorithm}

\subsection{Total Time Cost Considering Collaboration}
\label{5a}

The total time cost to finish all tasks can be calculated by Alg.~\ref{timecost}.
We define $\widetilde{\mathbb{T}}^{sort} = \{\sigma^{k}(m)\}_{\forall m\in \left[1:|\sigma^{k}|\right], k\in \mathcal{S}}$, in which all elements of $\widetilde{\mathbb{T}}^{sort}$ are sorted in increasing order of indexes $k$ and $m$.
In fact, it holds that $|\widetilde{\mathbb{T}}^{sort}|=|\sigma|$.
Each element $\sigma^{k}(m)\in \widetilde{\mathbb{T}}^{sort}$ can be treated as a generalized collaborative task since all tasks in the same $\sigma^{k}(m)$ have the same execution time, according to synchronization constraints in Sect.~\ref{3A}. 
The variable ${\rm delay}_{i}$ in Alg.~\ref{timecost} denotes the total wait time for robot~$i$ up to the execution of current $ct^{k}_{l}$ and is modified after each collaboration.
Given initial optimal run $\{\rho_i\}_{i\in \mathcal{N}}$, if there is no waiting time in previous collaborations, i.e., ${\rm delay}_i=0$, then the time robot $i$ arrives at the region of task $ct^{k}_{l}$ can be calculated by $t_{i}(ct^{k}_{l}) = \sum_{j = 1}^{j^k_l-1}\omega_{i}(\rho_{i}(j), \rho_{i}(j+1))$, $\forall~ct^{k}_{l}\in \widetilde{\mathbb{T}}_{i}$. 
Here $j^{k}_{l}$ denote the index of state $q$ in $\rho_{i}$ that satisfies $q = \rho_{i}(j^{k}_{l})$ and $q\in \mathcal{C}(ct^{k}_{l})$, for all $ct^{k}_{l}\in \widetilde{\mathbb{T}^{i}}$. The $t_{i}(\cdot)$ with subscript $i$ denotes that it is the ideal arrival time from robot $i$'s perspective without considering the possible waiting time in each collaboration.

Let $t(ct)$ denote the actual execution time of task $ct$, which is determined by the actual arrival time of the latest robot $\hat{i}$, i.e., $t(ct) = \max_{i\in \mathcal{R}(ct)}\left\{t_{i}(ct) + {\rm delay}_{i}\right\}$ (Line~3, Alg.~\ref{timecost}).
Note we reload the definition of $t_i(\cdot)$ as $t_{i}(\sigma^{k}(m)) = t_{i}(ct^{k}_{l})$, $\forall~ct^{k}_{l}\in \sigma^{k}(m) \cap \widetilde{\mathbb{T}}_{i}$.

The total time cost to complete all task execution plans can be calculated by $T^{\rm colla} = \sum_{i\in \mathcal{N}}T_{i}^{\rm colla} = \sum_{i\in \mathcal{N}}(T_{i}^{\rm indiv} + {\rm delay}_{i})$, where $T_{i}^{\rm indiv}=\sum_{j=1}^{|\rho_i|-1}\omega_{i}\left(\rho_i(j), \rho_i(j+1)\right)$, and ${\rm delay}_{i}$ is the total waiting time in all collaborations of $ct^{k}_{l}\in\widetilde{\mathbb{T}}_i$.
The variable ${\rm delay}_{i}$ may increase in each collaboration because each robot $i$ does not consider the collaboration with others in the local planning procedure.

\subsection{Greedy-based Execution Plan Adjusting}

The proposed execution plan adjusting mechanism (Alg.~\ref{ltpa}) tries to reduce the wait time in each collaboration, and sequentially traverses all collaborative task $ct\in \widetilde{\mathbb{T}}^{sort}$ according to their execution order, and performs the following two steps:

\textbf{Step 1:} 
The robots in $\mathcal{R}(ct)$ find the \textbf{latest} robot $\hat{i}$ to arrive at the region of $ct$ by inter-robot communication; and then the robot $\hat{i}$ investigates whether $T^{\rm colla}$ can be decreased by adjusting its pre-planned execution time of task $ct$ to an earlier time, to reduce the wait time of other robots in $\mathcal{R}(ct)$ (Lines 4-6).

\textbf{Step 2:} 
If Step $1$ fails, the robots then select the \textbf{earliest} robot $\hat{i}$ to execute task $ct$ in this step instead of the latest; and the robot $\hat{i}$ tries to postpone the planned execution time $t_{\hat{i}}(ct)$ to reduce the wait time for itself.

If no optimization happens in one iteration, i.e., $count = 0$, then Alg.~\ref{ltpa} will terminate (Lines 14-15). 
Otherwise, $count > 0$ means the modified task execution plan results in a smaller $T^{\rm colla}$ (Lines 7-8, 12-13), and the algorithm will go on to the next iteration to further optimize current plans.

Note that the robots only need to exchange information about when to task part in each collaborative tasks, so that their private information about individual tasks can be preserved.

\begin{algorithm}[!t]
\setstretch{0.8}
\SetKwInOut{Input}{Input}\SetKwInOut{Output}{Output}
\SetKwFunction{optimizeTime}{optimizeTime}
\label{ltpa}
\caption{ExecutionPlanAdjusting}
\small{
\Input{$\{\mathcal{P}_{i}\}_{i\in \mathcal{N}}$, $\{\rho_{i}\}_{i\in \mathcal{N}}$}
\While{${\rm true}$}{
    $count \leftarrow 0$\\
    \For{$ct\in \widetilde{\mathbb{T}}^{sort}$}{
    	 $ct^{prev}\leftarrow$ the previous task of $ct^k_l$ in $\widetilde{\mathbb{T}}_{i}$.\\
        $\hat{i} \leftarrow \arg \max\limits_{i\in \mathcal{R}(ct)}\left\{t(ct^{prev})-t_{i}(ct^{prev}) + t_{i}(ct^k_l)\right\}$\\
        $can\_opt \leftarrow{\rm OptTime}\left(\mathcal{P}_{\hat{i}}, ct^k_l, \rho_{\hat{i}}, isMax = {\rm true}\right)$\\
        \eIf{$can\_opt$}{
            $count \leftarrow count + 1$\\
        }
        {
            $\hat{i} \leftarrow \arg \min\limits_{i\in \mathcal{R}(ct)}\{t(ct^{prev})-t_{i}(ct^{prev}) + t_{i}(ct)\}$\\
            $can\_opt \leftarrow{\rm OptTime}\left(\mathcal{P}_{\hat{i}}, ct^{k}_{l}, \rho_{\hat{i}}, {\rm false}\right)$\\
            \If{$can\_opt$}{
                $count \leftarrow count + 1$\\
        }
        }
    }
    \If{$count = 0$}{
        break
    }
}
~\\
\DontPrintSemicolon
  \SetKwFunction{FMain}{\textbf{OptTime}}
  \SetKwProg{Fn}{Function}{:}{}
  \SetKwInOut{Input}{Input}\SetKwInOut{Output}{Output}
    \SetKwFunction{TimeCost}{TimeCost}
    \SetKwFunction{Dijkstra}{Dijkstra}
    \SetKwInOut{Return}{Return}
  \Fn{\FMain{$\mathcal{P}_{\hat{i}}$, $ct^k_l$, $\rho_{\hat{i}}$, $isMax$}}{
        \small{
            $ct^{prev}\leftarrow$ the previous task of $ct^k_l$ in $\widetilde{\mathbb{T}}_{i}$.\\
            \For{$q\in \mathcal{C}(ct)\backslash\{\rho_{\hat{i}}(j^{k}_{l})\}$}{
                $\rho'_{\hat{i}} \leftarrow$ \parbox[t]{.6\linewidth}{$\rho_{\hat{i}}[..j^{prev}-1]$\\
                +Dijkstra$\left(\mathcal{P}_{\hat{i}}, \rho_{\hat{i}}(j^{prev}), q\right)$\\         +Dijkstra$(\mathcal{P}_{\hat{i}}, q, \mathcal{Q}_{P}^{F})$}\\
                \If{\parbox[t]{1\linewidth}{$\left(isMax \bigwedge t(ct^{prev})-t_{i}(ct^{prev})+t'_{i}(ct^{k}_{l})<t_{i}(ct^{k}_{l})\right)$\\
                ${\rm or}$ $\left((isMax={\rm false})\bigwedge\right.$\\ $\left.t_i(ct^{k}_{l}) < t(ct^{prev})-t_{i}(ct^{prev})+t'_{i}(ct^{k}_{l})\le t(ct^{k}_{l})\right)$}}{
                    $T^{\rm colla}_{\rm cand} \leftarrow\ $TimeCost$(\rho'_{\hat{i}})$\\
                    \If{$T^{\rm colla}_{\rm cand}<T^{\rm colla}$}{
                        $\hat{\rho}_{i} \leftarrow \rho'_{\hat{i}}$\\
                        \KwRet true\\
                    }
                }
            }
            \KwRet false\\
            }
  }
}
\end{algorithm}

The detailed adjusting strategy is explained in Function ${\rm OptTime}$ of Alg.~\ref{ltpa}.
For the selected latest (or the earliest) arrival robot $\hat{i}$ in the collaboration of task $ct^{k}_{l}\in \widetilde{\mathbb{T}}^{\hat{i}}$, the set $\mathcal{C}(ct^{k}_{l})$ of $\mathcal{P}_{\hat{i}}$ actually provides all candidates states that can execute $ct^{k}_{l}$, namely, all possible arrival time to $ct^{k}_{l}$.
The robot $\hat{i}$ randomly traverses all state $q\in \mathcal{C}(ct^{k}_{l})\backslash\{\rho_{\hat{i}}(j^{k}_{l})\}$ until finding one candidate state $q$ which can optimize $T^{\rm colla}$.
For each candidate state $q$, the robot $\hat{i}$ searches a candidate run $\rho'_{\hat{i}}$ in $\mathcal{P}_{\hat{i}}$. The candidate run $\rho'_{\hat{i}}$ keeps $\rho_{\hat{i}}[..j^{prev}-1]$ unchanged, but finds subsequent sequence from $\rho_{\hat{i}}(j^{prev})$ to the accepting states of $\mathcal{P}_{\hat{i}}$, forcing it to pass through $q$ (Line 20). Here $\rho_{\hat{i}}(j^{prev})$ is the collaborative state of task $ct^{prev}$ in $\rho_{\hat{i}}$ and $ct^{prev}$ is the previous collaborative task of $ct^{k}_{l}$ in $\widetilde{\mathbb{T}}^{\hat{i}}$.
The candidate run $\rho'_{\hat{i}}$ is selected to be the new $\rho_{\hat{i}}$ if the following two conditions hold (Lines~21-25): 
\begin{enumerate}
    \item if robot $\hat{i}$ is the latest robot, it holds that $t(ct^{prev})-t_{i}(ct^{prev})+t'_{i}(ct^{k}_{l})<t_{i}(ct^{k}_{l})$; 
    
    if robot $\hat{i}$ is the earliest robot, it holds that $t_i(ct^{k}_{l}) < t(ct^{prev})-t_{i}(ct^{prev})+t'_{i}(ct^{k}_{l})\le t(ct^{k}_{l})$.
    \item $T^{\rm colla}_{\rm cand}<T^{\rm colla}$.
\end{enumerate}
The condition $1)$ ensures that in the candidate run $\rho'_{\hat{i}}$, the time when the latest robot $\hat{i}$ arrives $ct^{k}_{l}$ is advanced (or delayed for the earliest robot), such that the waiting time in the collaboration of task $ct^{k}_{l}$ is reduced. 
Here $t'_i(ct^{k}_{l})$ is calculated according to the candidate run $\rho'_{\hat{i}}$.
The condition $2)$ further guarantees the above local greedily adjustment procedure will also contribute to the total time cost $T^{\rm colla}$.
If the algorithm cannot find a qualified candidate run $\rho'_{\hat{i}}$ that satisfies the above two conditions, then returns false.

\subsection{Monotonicity, Complexity, and Optimality}

\begin{proposition}
\label{prop1}
Algorithm~\ref{ltpa} is monotonic and terminates within finite iterations. 
\end{proposition}

\begin{Proof}
Given the collaborative task assignments, $T^{\rm colla}$ monotonically decreases from $T^{\rm colla}_{\rm init}$ to $T^{\rm indiv}$, because it decreases to a strictly smaller $T^{\rm colla}_{\rm cand}$ in each iteration (Lines 22-25, Alg.~\ref{ltpa}) until the algorithm terminates. Here $T^{\rm indiv} = \sum_{i\in \mathcal{N}} T_{i}^{\rm indiv}$, and $T^{\rm colla}_{\rm init}$ is the time cost of initial execution plans for robots in $\mathcal{N}$.
Considering the resolution of discretization of the environment is limited, we conclude that Algorithm~\ref{ltpa} will terminate within finite iterations.
\hfill~$\blacksquare$
\end{Proof}

\begin{proposition}
The worst case time complexity of Algorithm~\ref{ltpa} is $\mathcal{O}\left((T^{\rm colla}_{\rm init}-T^{\rm indiv})\cdot |\mathcal{P}_i|\cdot(E\cdot \lg E + |\sigma|\cdot N)\right)$, where $N$ is the number of robots, $|\mathcal{P}_{i}|$ and $E$ are the maximum number of states and edges in all $\mathcal{P}_{i}$ respectively. 
$|\sigma|$ is the length the selected accepting run in Sect.~\ref{3A}.
\end{proposition}

\begin{Proof}
In the worst case, Alg.~\ref{ltpa} will traverse all candidate collaborative states in $\mathcal{P}_i$ in each cycle of the while loop (Lines 1-15). 
The number of cycles of the while loop is limited by $\mathcal{O}(T^{\rm colla}_{\rm init}-T^{\rm indiv})$ as in Proposition~\ref{prop1}.
For each candidate collaborative state, the function ${\rm OptTime}$ utilizes Dijkstra algorithm to find a run $\rho'_i$ with complexity $\mathcal{O}(E\cdot \lg E)$, and also calls Alg.~\ref{timecost} to validate the candidate run with time complexity $|\sigma|\cdot N$.
To summarize, the time complexity of Alg.~\ref{ltpa} is $\mathcal{O}\left((T^{\rm colla}_{\rm init}-T^{\rm indiv})\cdot |\mathcal{P}_i|\cdot(E\cdot \lg E + |\sigma|\cdot N)\right)$.

\hfill~$\blacksquare$
\end{Proof}

\textbf{Optimality: } The proposed framework does not guarantee to obtain the optimal execution plans, because Algorithm~\ref{ltpa} utilizes a greedy strategy rather than exhaustively investigating the combinations of all possible feasible execution plans. Given assignments of collaborative tasks, a dynamic programming method can be used to find the optimal execution plans for the robots, but it has exponential complexity in each step.
Moreover, the choice of the selection of collaborative task sequence $\sigma$ in Sect.~\ref{3A} also influences the performance of the final results.

\section{Simulation Results}
\label{simulation}

In this section, we evaluate the scalability of the proposed framework comparing with a baseline method based on global product automaton, and showcase the effectiveness of the proposed
execution plan adjusting mechanism
in reducing the total time cost to finish all tasks. 
All the experiments are performed on a Ubuntu 16.04 server with CPU Intel Xeon E5-2660 and 128 GB of RAM. We use Z3~\cite{10.1007/978-3-540-78800-3_24} SMT solver to solve the SMT model in the task allocation procedure. 

We assume a grid map environment where individual tasks and collaborative tasks are randomly distributed. Each robot $i$ has a local task specification $\varphi_{i}$, e.g., $\varphi_{i} = (\Diamond \pi^{ts_{1}}) \wedge (\Diamond \pi^{ts_{2}}) \wedge (\Diamond \pi^{ts_{3}}) \wedge (\Diamond \pi^{ts_{4}})\wedge (\neg \pi^{ts_{1}}~{\rm U}~\pi^{ts_{4}})$. 
Additionally, the team of robots needs to collaborate to satisfy collaborative task specification $\phi$, e.g., $\phi = (\Diamond \pi^{ct_{1}}) \wedge (\Diamond \pi^{ct_{2}}) \wedge (\Diamond \pi^{ct_{4}}) \wedge (\neg \pi^{ct_{3}}~{\rm U}~\pi^{ct_{2}}) \wedge (\Diamond (\pi^{ct_{4}} \wedge (\Diamond \pi^{ct_{3}})))$ , where each $ct\in \widetilde{\mathbb{T}}$ requires collaboration of several types of robots having capabilities from $\{c_{1}, c_{2}, c_{3}\}$.
The type and amount of robots needed for each collaborative task are randomly generated in the experiments.
The final execution plans in an experiment with $3$ robots are depicted in Fig.~\ref{environment}.

\begin{table}[]
\footnotesize
\setlength{\abovecaptionskip}{0.1cm}
\caption{Baseline Method (B.S.) Vs The proposed Method (Proposed)}
\centering
\label{compare}
\begin{tabular}{|c|c|c|c|c|p{0.6cm}<{\centering}|}
\hline
Env.                      & $N$ & $|\dot{\mathcal{P}}|$ & B.S. & Proposed & Speed\\
 Size   & & & (sec/sec) &(sec/sec) & Up\\
\hline
\multirow{3}{*}{$5\times 5$}   & $2$          &   $2.22\times 10^{3}$      & $30/0.58$    &  $30/0.11$    &  $5$   \\ \cline{2-6} 
                               & $3$          &   $7.17\times 10^{4}$       &  $57/84.62$    &     $51/0.94$  &  $90$  \\ \cline{2-6} 
                               & $5$          &   $8.20\times 10^{7}$       &   -/-  &  $59/7.91$   &  -   \\ \hline
\multirow{3}{*}{$10\times 10$} & $2$          &  $4.08\times 10^{4}$      & $40/16.15$     &     $39/0.56$   & $29$  \\ \cline{2-6} 
                               & $3$          &  $4.13\times 10^{6}$       &  $78/104571$    &    $78/3.57$    &  $29292$ \\ \cline{2-6} 
                               & $5$          &   $4.87\times 10^{10}$      &    -/-  &   $148/32.00$     &  - \\ \hline
\multirow{3}{*}{$15\times 15$} & $2 $         &   $2.04\times 10^{5}$       &   $88/117.42$   & $78/1.23$    &  $95$    \\ \cline{2-6} 
                               & $3 $         &   $4.80\times 10^{7}$       &   -/-   &   $145/8.45$    &  -  \\ \cline{2-6} 
                               & $5$          &   $2.52\times 10^{12}$       &   -/-   &   $163/102.60$   &   - \\ \hline
\end{tabular}
\vspace{0.3ex}
{\justifying \\
The items in column B.S. and Proposed represent:  $T^{\rm colla}/t^{\rm cal}$, where $T^{\rm colla}$ is the total time cost to satisfy all task specifications and $t^{\rm cal}$ is the solving time. 
The ``-'' items indicate memory overflow. \par}
\vspace{1ex}
\end{table}

 $\textbf{(1) Scalability and Efficiency.}$ 
 To evaluate the scalability of the proposed method, we compare the proposed method with a baseline method based on the product automaton denoted by $\dot{\mathcal{P}}$ with state pruning techniques, whose construction process is similar to \cite{tumovaDecompositionMultiagentPlanning2015}. 
The details are omitted here due to space limitations.

We investigate the performance of the two methods in different environment sizes and robot numbers, as shown in Table~\ref{compare}.
The proposed method quickly solves all the cases and is much faster in comparison with the baseline method.
The total time cost of the solutions provided by the proposed method is better than that of the baseline method, since there is no efficient way to find the optimal run w.r.t the total time cost (which takes into account the wait time in each collaboration) in the product automaton $\dot{\mathcal{P}}$. 
The results illustrate that the proposed method can provide high-quality solutions to Problem $1$, and also has high scalability comparing with the baseline method.

\begin{figure}[t]
	\centering
	\begin{overpic}[width=\linewidth]{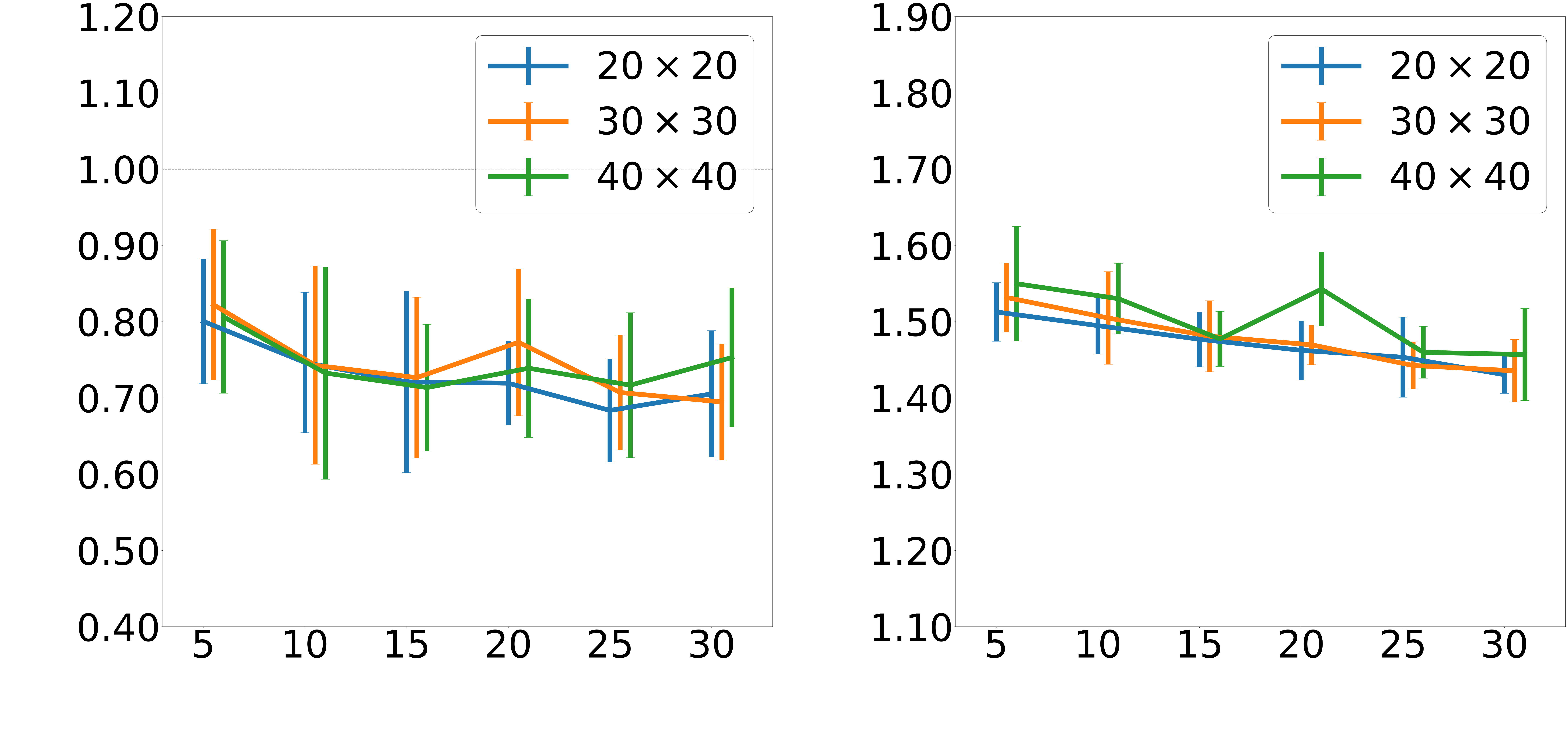}
    \put(1,18){\rotatebox{90}{\scriptsize{$T^{\rm colla}/T^{\rm colla}_{\rm init}$}}}
    \put(51,18){\rotatebox{90}{\scriptsize{$t^{\rm cal}/t^{\rm cal}_{\rm -adj}$}}}
    \put(27,2){\scriptsize{$N$}}
    \put(78,2){\scriptsize{$N$}}
	\put(27,-1){\scriptsize{(a)}}
	\put(78,-1){\scriptsize{(b)}}
	\end{overpic}
	\caption{The comparison of the framework with the execution plan adjusting mechanism versus without it: (a) $T^{\rm colla}/T^{\rm colla}_{\rm init}$; (b) $t^{\rm cal}/t^{\rm cal}_{\rm -adj}$.}
	\label{ratio}
\end{figure}

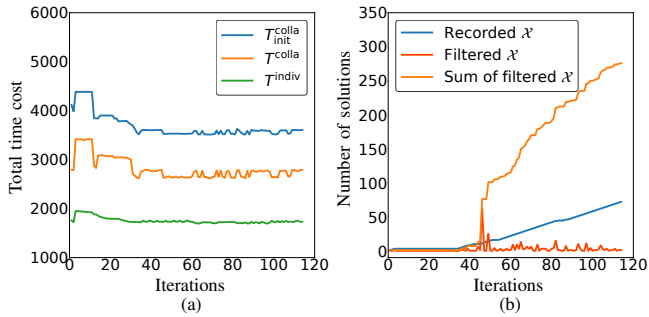
\begin{figure}[t]
	\centering
	\begin{overpic}[width=\linewidth]{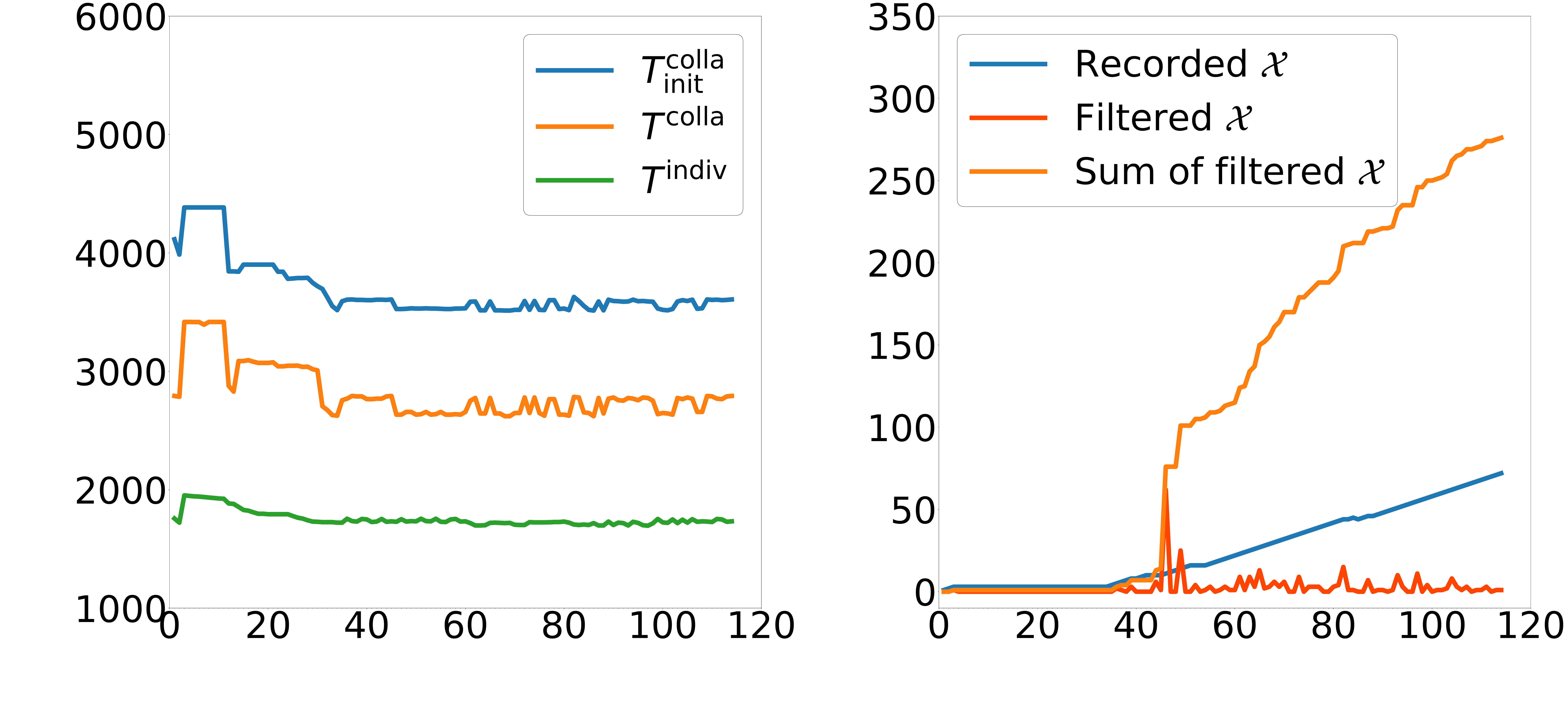}
	\put(24,2){\scriptsize{Iterations}}
	\put(73,2){\scriptsize{Iterations}}
	\put(1.5,17){\rotatebox{90}{\scriptsize{Total time cost}}}
	\put(52,15){\rotatebox{90}{\scriptsize{Number of solutions}}}
	\put(28,-1){\scriptsize{(a)}}
	\put(77,-1){\scriptsize{(b)}}
	\end{overpic}
	\caption{(a) The optimization process of the total time cost and (b) the filtered SMT solutions in each iteration by utilizing the filtering strategy in an experiment with $30$ robots.}
	\label{one-case}
\end{figure}
$\textbf{(2) Optimization\ Performance.}$
To investigate the optimization performance of the execution plan adjusting mechanism (Alg.~\ref{ltpa}), we conduct extensive simulations under various numbers of the robots and environment sizes. 
The number of robots varies from $\{5, 10, 15, 20,25,30\}$ and three types of environment sizes are investigated. 
In each setting, we randomly distribute the tasks in the grid map environment for $10$ independent trials under the same task specifications. 
We terminate the solving process when the running time exceeds $30$ minutes, and choose the best result up to the end time. 
As shown in Fig.~\ref{ratio}, the proposed execution plan adjusting mechanism can reduce $T^{\rm colla}$ to about $70\%\sim 80\%$ of its initial value $T^{\rm colla}_{\rm init}$, while $t^{\rm cal}$, the running time  of the framework with the execution plan adjusting mechanism, is $1.x$-proportional to $t^{\rm cal}_{\rm -adj}$ the running time of the proposed framework without the execution plan adjusting procedure. 
This indicates that the proposed adjusting mechanism can trade a small increase of running time for a large improvement of performance, and the conclusion holds under different sizes of robot teams. 
We have attached a video to illustrate the improvement of the obtained execution plans caused by the proposed adjusting mechanism.

As shown in Fig.~\ref{one-case}(a), the total time cost $T^{\rm colla}$ after optimization is approaching $T^{\rm indiv}$ after execution plan adjusting procedure in each iteration, i.e., under each task assignment scheme, in an experiment with 30 robots in a $20\times 20$ grid map.
It illustrates that the execution plan adjusting mechanism can greatly optimize the initial task execution plans in each iteration.
Note that $T^{\rm indiv}$ is not the lower bound of the optimal solution to Problem $1$.
In addition, as shown in Fig.~\ref{one-case}(b), by utilizing the filtering strategy proposed in Sect.~\ref{3B}, many obtained feasible solutions of the task assignment problem can be filtered to prevent unnecessary computation in the same experiment as in Fig.~\ref{one-case}(a).  
The strategy only needs to maintain a set of solutions with its size propositional to the number of iterations, 
Note that the filtering efficiency depends on the inner computation mechanism of the SMT solver and the specific task requirements. 

\section{CONCLUSIONS AND FUTURE WORK}
We propose a hierarchical multi-robot temporal task planning framework in which robots can efficiently synthesize execution plans that satisfy both individual and collaborative temporal logic task specifications, while also protecting their private information about individual tasks.
The expected actual performance of execution plans can be iteratively improved by reducing the waiting time for the robots in each collaboration, utilizing the proposed adjusting mechanism.
Simulation results illustrate the correctness and efficiency of the proposed method.
Future work is to improve the efficiency of the task allocation procedure and expand the current framework into complete LTL.

\section*{ACKNOWLEDGMENT}
The work was supported by Natural Science Foundation of China under Grant 61873235, Zhejiang Provincial Natural Science Foundation of China under Grant LZ19F030002, and the NSFC-Zhejiang Joint Fund for the Integration of Industrialization and Informatization under Grant U1909206. 
This work was also supported by the NSFC-Zhejiang Joint Fund for the Integration of Industrialization and Informatization under Grants U1709203, U1809212 and the Key Research and Development Program of Zhejiang Province under Grant 2019C03109.




\bibliographystyle{ieeetr} 
\bibliography{root} 

\end{document}